\documentclass[preprint,3p,times,fleqn]{elsarticle}
\usepackage{amsmath}
\usepackage{amssymb}
\usepackage{psfrag}
\usepackage{color}
\usepackage{stfloats}
\usepackage{mathrsfs}
\usepackage{tabularx}
\usepackage{booktabs}
\usepackage{colortbl}
\usepackage{rotating}
\usepackage{boldline}
\usepackage{bm}
\usepackage{changes}
\usepackage{url}
\DeclareMathOperator*{\argmin}{arg\,min}
\DeclareMathOperator*{\argmax}{arg\,max}
\usepackage{siunitx}

\usepackage{lineno}

\usepackage{multirow}

\journal{arxiv}
\begin{document}
\renewcommand{\topfraction}{0.98}	
\renewcommand{\bottomfraction}{0.98}
\setcounter{topnumber}{3}
\setcounter{bottomnumber}{3}
\setcounter{totalnumber}{4}         
\setcounter{dbltopnumber}{4}        
\renewcommand{\dbltopfraction}{0.98}	
\renewcommand{\textfraction}{0.05}	
\renewcommand{\floatpagefraction}{0.5}	
\renewcommand{\dblfloatpagefraction}{0.5}	
\newcommand{\beq}{\begin{equation}}
\newcommand{\eeq}{\end{equation}}
\newcommand{\divg}{\mbox{\rm{div}}\,}
\newcommand{\Divg}{\mbox{\rm{Div}}\,}
\newcommand{\D}  {\displaystyle}
\newcommand{\DS} {\displaystyle}
\newcommand{\RM}[1]{\textit{\MakeUppercase{\romannumeral #1{}}}}
\newtheorem{remark}{\bf{{Remark}}}
\def\sca   #1{\mbox{\rm{#1}}{}}
\def\mat   #1{\mbox{\bf #1}{}}
\def\vec   #1{\mbox{\boldmath $#1$}{}}
\def\scas  #1{\mbox{{\scriptsize{${\rm{#1}}$}}}{}}
\def\scaf  #1{\mbox{{\tiny{${\rm{#1}}$}}}{}}
\def\vecs  #1{\mbox{\boldmath{\scriptsize{$#1$}}}{}}
\def\tens  #1{\mbox{\boldmath{\scriptsize{$#1$}}}{}}
\def\tenf  #1{\mbox{{\sffamily{\bfseries {#1}}}}}
\def\ten   #1{\mbox{\boldmath $#1$}{}}
\def\Ass  {\overset{\hspace*{0.4cm} n_{\scas{el}}}
          {\underset{\scaf{c},\scaf{d}=1}{\msf{A}}}}
\def\ltr   #1{\mbox{\sf{#1}}}
\def\bltr  #1{\mbox{\sffamily{\bfseries{{#1}}}}}
\sloppy
\begin{frontmatter}
\title{\Large Probabilistic learning of the Purkinje network from the electrocardiogram}


\author[01]{Felipe \'{A}lvarez-Barrientos}
\author[06]{Mariana Salinas-Camus}

\author[04,05]{Simone Pezzuto}

\author[01,02,03]{Francisco~Sahli Costabal\corref{cor1}}
\ead{fsc@ing.puc.cl}

\cortext[cor1]{Corresponding author}
\address[01]{Department of Mechanical and Metallurgical Engineering, School of Engineering, Pontificia Universidad Cat\'olica de Chile, Santiago, Chile}

\address[06]{Intelligent Sustainable Prognostics Group, Aerospace Structures and Materials Department, Faculty of Aerospace Engineering, Delft University of Technology, Delft, Netherlands}

\address[04]{Laboratory of Mathematics for Biology and Medicine, Department of Mathematics, Università di Trento, Trento, Italy}

\address[05]{Center for Computational Medicine in Cardiology, Euler Institute, Università della Svizzera italiana, Lugano, Switzerland}

\address[02]{Institute for Biological and Medical Engineering, Schools of Engineering, Medicine and Biological Sciences, Pontificia Universidad Cat\'olica de Chile, Santiago, Chile}

\address[03]{Millennium Institute for Intelligent Healthcare Engineering, iHEALTH}

\begin{abstract} %

The identification of the Purkinje conduction system in the heart is a challenging task, yet essential for a correct definition of cardiac digital twins for precision cardiology. Here, we propose a probabilistic approach for identifying the Purkinje network from non-invasive clinical data such as the standard electrocardiogram (ECG). We use cardiac imaging to build an anatomically accurate model of the ventricles; we algorithmically generate a rule-based Purkinje network tailored to the anatomy; we simulate physiological electrocardiograms with a fast model; we identify the geometrical and electrical parameters of the Purkinje-ECG model with Bayesian optimization and approximate Bayesian computation. The proposed approach is inherently probabilistic and generates a population of plausible Purkinje networks, all fitting the ECG within a given tolerance. In this way, we can estimate the uncertainty of the parameters, thus providing reliable predictions. We test our methodology in physiological and pathological scenarios, showing that we are able to accurately recover the ECG with our model. We propagate the uncertainty in the Purkinje network parameters in a simulation of conduction system pacing therapy. Our methodology is a step forward in creation of digital twins from non-invasive data in precision medicine. An open source implementation can be found at \url{http://github.com/fsahli/purkinje-learning}

\end{abstract}
\begin{keyword}
Purkinje Network; Machine learning; Bayesian inference; Approximate Bayesian Computation; Digital Twin; Cardiac electrophysiology.
\end{keyword}
\end{frontmatter}


\section{Motivation}\label{intro}


Cardiovascular diseases are the leading causes of death in the world, taking around 19 million lives in 2020~\cite{tsao2023heart}. Arrhythmias, which are abnormal heart rhythms, can lead to the impairment of cardiac function. 
One of the main components of the cardiac electrical system is the Purkinje network, located in the subendocardium of the ventricles. Purkinje cells are larger than cardiomyocytes and have fewer myofibrils and more mitochondria. The electrical propagation of the excitation wave in Purkinje cells is faster than in other cardiac myocytes. A functional Purkinje network is essential to sync the activations and contractions of the left and right ventricles and maintain a consistent cardiac rhythm~\cite{dubin1996rapid}. Purkinje fibers were first documented more than a century ago~\cite{tawara1906reizleitungssystem}; yet, to date, there is no technique to fully reconstruct their geometry \textit{in vivo} in humans~\cite{Cetingul2011,magat20213d,goodyer2022vivo,tsao2023heart}. Due to these limitations, there have been multiple attempts to represent the Purkinje network with simplified models from a functional and geometric perspective.

From a modeling point of view, the Purkinje network is oftentimes surrogated by a fast-conducting subendocardial layer with a limited number of discrete early activation sites~\cite{Pezzuto2017Fast}. Thanks to its small number of parameters, the surrogate model is computationally efficient and enables fast model personalization from clinical data~\cite{pezzuto2021reconstruction,gillette_framework_2021, GeodesicBP2023}. However, the surrogate model cannot capture the complexity of the ventricular activation under pathological conditions such as a bundle branch block. Recently, algorithms based on fractal trees~\cite{Lindenmayer1968} have emerged as a tool for automatic network generation. Current models consist of curved branches that grow on the endocardial surface \cite{Ijiri2008, SahliCostabal2015,gillette2021automated}. Other models are generated with hierarchical networks that add sub-sequentially smaller fibers \cite{Sebastian2011, Sebastian2013}.

Regardless of the representation, there is an increasing interest in creating a digital twin of the Purkinje network from available clinical data. Considering the lack of imaging modalities available for the Purkinje network, there have been attempts to recover the structure of the Purkinje network from electrical measurements. Early studies focused on invasive electroanatomical mapping data, where activation times inside the ventricles are known~\cite{cardenes2015estimation, palamara2014computational, Vergara2014}. More recent studies propose a rule-based method to non-invasively reconstruct the patient-specific Purkinje network from the electrocardiogram (ECG)~\cite{gillette2021automated}.

Purkinje network identification from the ECG is an ill-posed inverse problem. First, the number of parameters describing the network is high, with several of them only marginally affecting the ECG. Thus, the set of plausible Purkinje networks explaining a given ECG is large. Second, depending on the condition of the patient, parts of the Purkinje network may be blocked and not influence the ventricular activation at all. Here, identification of the tree structure from the ECG is not possible. From a clinical perspective, however, partially blocked parts may be the target of therapeutic options, for instance, conduction system pacing~\cite{ellenbogen2023evolving}. For these reasons, it is fundamental to quantify what is and is not possible to infer the Purkinje network from the standard ECG.

\begin{figure}[t]
	\centering
	\includegraphics[width=\textwidth]{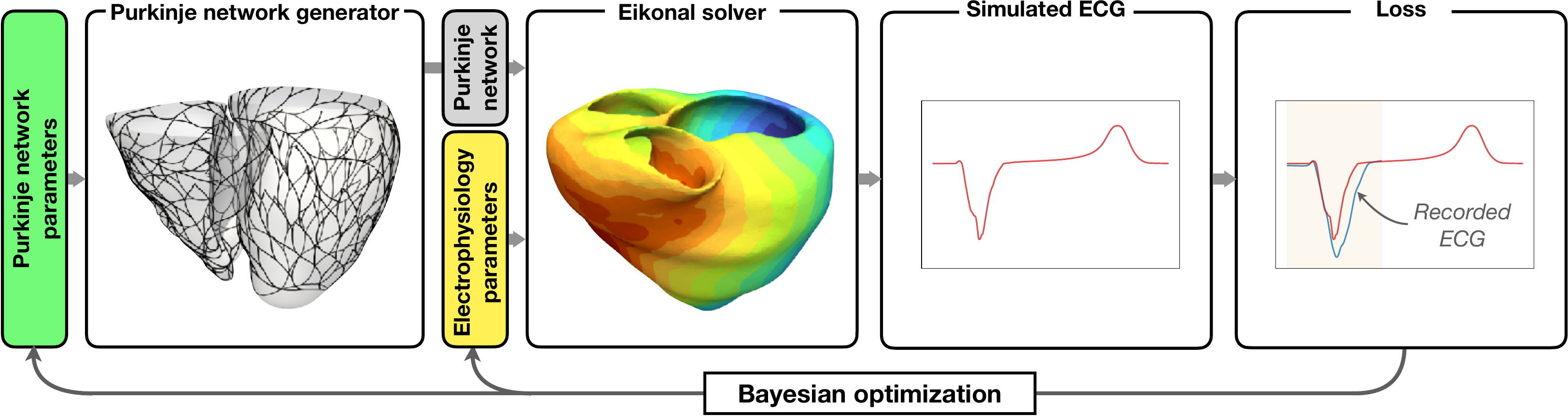}
	\caption{\textbf{Process outline.} We first create a parametric Purkinje network on the left and right ventricles. For a given Purkinje network and electrophysiology parameters we simulate activation times with the Eikonal equation, from which we can compute an ECG. We compare this ECG with the patient data and attempt to find the optimal parameters with Bayesian optimization.}
	\label{fig:outline}
\end{figure}

This work addresses the Purkinje network identification problem with a probabilistic approach. In other words, we aim to recover a \textit{distribution} of Purkinje networks that can explain the observed ECG rather than providing a single estimate. Within a probabilistic framework, we can quantify the uncertainty in the estimation process and assess predictions with confidence intervals. We tackle this challenging problem by integrating advanced techniques that include electrophysiology modeling, machine learning, and probability density estimation. Figure~\ref{fig:outline} illustrates the first part of the process. We start by creating a parametric representation of the Purkinje network using fractal trees~\cite{SahliCostabal2015}. Then, we couple the Purkinje network with the myocardium to simulate the ECG using a fast solver based on the eikonal equation~\cite{Pezzuto2017Fast}. We find the network parameters that best represent the patient's ECG by defining a loss function that compares the predicted ECG by our model and the observed data~\cite{pezzuto2021reconstruction,PezzutoBayes2022}. We minimize the loss function using Bayesian optimization, a machine-learning technique where the objective function is approximated by a Gaussian process~\cite{garnett2023bayesian}. Once we have a set of parameters that best fits the data, we estimate the distribution of parameters that explain the data by using the Gaussian process of the loss function as a prior distribution for approximate Bayesian computation~\cite{sunnaaker2013approximate}. Therefore, we can draw samples from the posterior distribution of parameters.

This manuscript is organized as follows: in Section~\ref{sec:methods} we present the methodology to generate the Purkinje network, the forward model to compute the ECG for a given Purkinje network, and the probabilistic pipeline to identify a distribution of trees. In Section~\ref{sec:results}, we show the results of the proposed method in a synthetic example and on 4 patients with different conditions. We finalize this Section by showing the predictive capabilities with quantified uncertainty of our method. We finish this manuscript with a discussion and outlook in Section~\ref{sec:discussion}.

\section{Methods}
\label{sec:methods}

\subsection{Purkinje network generation}
\begin{figure}[t]
	\centering
	\includegraphics[width=0.9\textwidth]{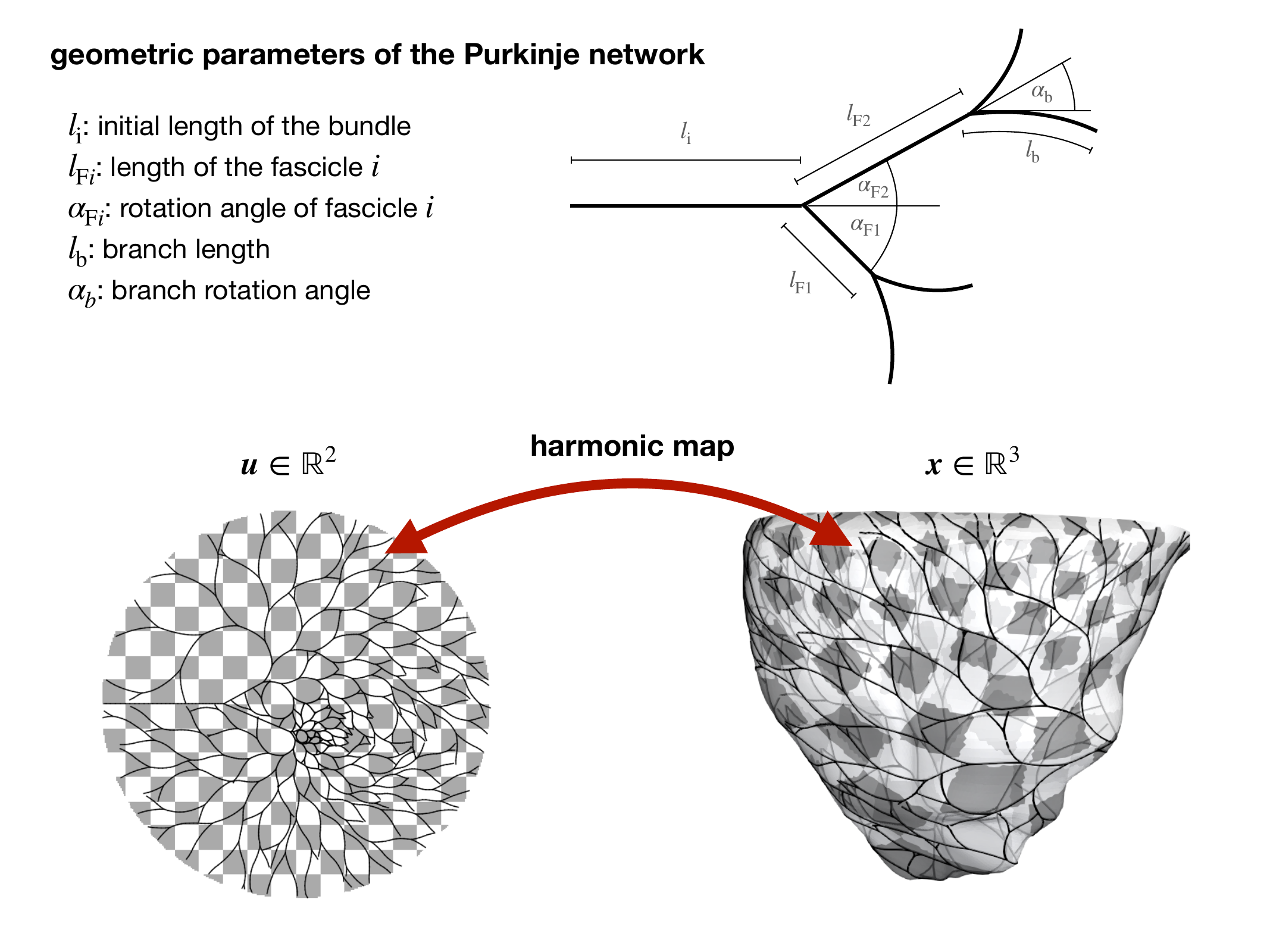}
	\caption{\textbf{Purkinje network generation.} Top, we show the geometric parameters that determine the structure of the Purkinje network. Bottom, we use a harmonic map to grow the Purkinje network in 2D, which is later mapped to the endocardial surface.}
	\label{fig:purkinje}
\end{figure}
We model the Purkinje network as fractal tree, in a similar way as in our previous work \cite{SahliCostabal2015}. We select a root node and initial direction and length $l_{\rm i}$ to represent the bundles. Then, we create an arbitrary number of branches, that we call fascicles with different branching angles $\alpha_{\rm{F}i}$ and lengths $l_{\rm{F}i}$. From the end of each fascicle we start growing the fractal tree. The branches are characterized by 3 parameters that are common for the entire tree: the branch length $l_{\rm b}$, the branching angle $\alpha_{\rm b}$ and the repulsion parameter $w$. Each branch is discretized a number of segments $N_s$ which are added simultaneously to end of each growing branch according to the following rule:
\begin{align}
    \vec{d}^{i+1} &= \frac{\vec{d}^i + w \nabla d_{\rm CP}(\vec{x}^i)}{\|\vec{d}^i + w \nabla d_{\rm CP}(\vec{x}^i)\|}, \quad i > 1, \\
    \vec{x}^{i+1} &= \vec{x}^i + \frac{l_{\rm b}}{N_s}\vec{d}^{i+1},
\end{align}
where $\vec{d}^{i+1}$ represents the normalized direction of the $(i+1)$-th segment, which is comprised of end point $\vec{x}^i$, $\vec{x}^{i+1}$. The function $d_{\rm CP}(\vec{x})$ represents the distance to the closest point already in the tree. Thus, the gradient $\nabla d_{\rm CP}$ provides a repulsive direction from the existing points. The initial direction $\vec{d}^1$ of each branch is determined by a rotation of either $+\alpha_{\rm b}$ or $-\alpha_{\rm b}$ with respect to the direction of the last segment from the parent branch. The main geometric parameters of the Purkinje tree are shown in Figure~\ref{fig:purkinje}, top row.

To avoid the constant projection of new points to the 3D surface, we flatten the ventricular geometry. In particular, we map the endocardial surface with coordinates $\vec{x} \in \mathbb{R}^3$ to a unit disk with coordinates $\vec{u} \in \mathbb{R}^2$ using harmonic maps \cite{remacle2010high}, as shown in Figure~\ref{fig:purkinje}, bottom row. These maps are quasi-conformal, meaning that angles are approximately preserved. Then, if we draw a straight line in the unit disk, it should remain a straight line in the ventricular geometry. The area is not preserved and we need take into account a scaling factor $s(\vec{u})$ for the length, such that $\|d\vec{x}\| = s(\vec{u})\|d\vec{u}\|$. To compute the harmonic map, we only need to solve the Laplace equation twice on the surface, which we achieve using finite elements. 

\subsection{Forward ECG model}

We model cardiac activation in the Purkinje tree and in the myocardium with the eikonal equation~\cite{Pezzuto2017Fast}. The eikonal equation models the spread of the activation front given a (possibly anisotropic) conduction velocity and initiation sites. For the sake of simplicity, we introduce the forward model and the notation for a single Purkinje tree. Here, we denote a Purkinje tree by $\mathcal{P} = \bigcup_{k=1}^{N} \Gamma_k$, where each branch $\Gamma_k$ is a smooth open curve in $\mathbb{R}^3$. The boundary $\partial\mathcal{P} = \bigl\{ \vec{p}_i \bigr\}_{i=0}^M$ contains the nodes of the tree, denoted by $\vec{p}_i$. We assume that the nodes are ordered as follows: $\vec{p}_0$ is the root node; for $i=1,\ldots,M_t$, $\vec{p}_i$ is a terminal node or Purkinje-Myocardium Junctions (PMJs); for $i=M_t+1,\ldots,M$, $\vec{p}_i$ is a branching point, that is the intersection of exactly 3 different branches. We represent the active myocardium by a smooth domain $\Omega\subset\mathbb{R}^3$. The PMJs are by definition the points of contact between the Purkinje tree and the myocardium, and specifically the endocardium. That is, $\vec{p}_i\in\partial\Omega$, $i=1,\ldots,M_t$.

The activation in the tree, denoted by $\tau^{\mathcal{P}}\colon \bar{\mathcal{P}}\to\mathbb{R}$, measures the travel time from the root node to the rest of the tree. Similarly, $\tau^{\mathcal{M}}\colon \Omega\to\mathbb{R}$ is the activation map in the myocardium. They solve the following systems:
\begin{equation}
\label{eq:eikotree}
\left\{\begin{aligned}
& c^{\mathcal{P}}(\vec{x}) | \nabla \tau^{\mathcal{P}}(\vec{x})\cdot \vec{t}(\vec{x})| = 1, && \vec{x}\in \mathcal{P}, \\
& \tau^{\mathcal{P}}(\vec{p}_0) = t_0, & \\
& \tau^{\mathcal{P}}(\vec{p}_i) = \min \bigl\{ \tau^{\mathcal{M}}(\vec{p}_i), \tau^{\mathcal{P}}(\vec{p}_i) \bigr\}, && \vec{p}_i \in \partial{P}, i=1,\ldots,M_t, \\
& \tau^{\mathcal{P}}(\vec{p}_i) = \min \bigl\{ \tau^{\mathcal{P}}|_{\partial\mathcal{P}}(\vec{p}_i) \bigr\}, && i = M_t+1,\ldots,M,
\end{aligned}\right.
\end{equation}
for the Purkinje activation, where $c^{\mathcal{P}}\colon\mathcal{P}\to\mathbb{R}$ and $\vec{t}$ are respectively the conductivity velocity field and the tangent field in the Purkinje tree, and
\begin{equation}
\label{eq:eikomyo}
\left\{\begin{aligned}
&\sqrt{\ten{D}(\vec{x})\nabla\tau^{\mathcal{M}}(\vec{x})\cdot\nabla\tau^{\mathcal{M}}(\vec{x})} = 1, && \vec{x}\in\Omega, \\
&\tau^{\mathcal{M}}(\vec{p}_i) = \min \bigl\{ \tau^{\mathcal{M}}(\vec{p}_i), \tau^{\mathcal{P}}(\vec{p}_i) \bigr\}, && \vec{p}_i \in \partial{P}, i=1,\ldots,M_t, \\
\end{aligned}\right.
\end{equation}
in the myocardium, where $\ten{D}^{\mathcal{M}}\colon\Omega\to\mathbb{R}^{3\times 3}$ is the conductivity tensor field, assumed symmetric, positive-definite. In particular, in the myocardium $\ten{D}$ is anisotropic and proportional to the monodomain conductivity tensor
\begin{align}
\ten{D}(\vec{x})  &= \alpha^2 \ten{G}_i(\ten{G}_i+\ten{G}_e)^{-1}\ten{G}_e, \\
\ten{G}_{\{i,e\}} &= \sigma^{\{i,e\}}_t \mathbf{I} + \bigl(\sigma^{\{i,e\}}_l - \sigma^{\{i,e\}}_t \bigr)\vec{f}(\vec{x})\otimes\vec{f}(\vec{x}),
\end{align}
where $\ten{G}_i$ and $\ten{G}_e$ respectively the intracellular and extracellular conductivity tensor, $\alpha$ is a scaling constant, $\vec{f}(\vec{x})$ is the fiber field, and $\sigma^{\{i,e\}}_{\{l,t\}}$ are the intra-/extra-cellular conductivities longitudinal and orthogonal to the fibers. 

Note that the trace $\tau^{\mathcal{P}}|_{\partial\mathcal{P}}$ in \eqref{eq:eikotree} evaluated on a branching point is multi-valued: the last forces it to be equal to the minimum activation time across all branches. The equations have a simple interpretation: within each branch or in the myocardium, $\tau(\vec{x})^{\mathcal{P}}$ and $\tau(\vec{x})^{\mathcal{M}}$ solve the eikonal equation. The activation at the PMJs is the earliest between those arriving either from the tree or from the myocardium. In other words, the coupling is bidirectional: either the electric signal from the tree activates the myocardium (orthodromic or normal activation), or the electric signals enters into the tree from the myocardium (antidromic activation). Note that we do not consider here activation delays at the PMJs, as it normally would happen due to a source-sink mismatch.

The coupled problem \eqref{eq:eikotree}-\eqref{eq:eikomyo} is well-posed and has a unique viscosity solution for every choice of the initial root time $t_0$. To see this, we note that the solution in the myocardium can be written as:
\begin{equation}
\tau^{\mathcal{M}}(\vec{x}) = \min_{i=1,\ldots,M_t} \Bigl\{  \delta^\mathcal{M}(\vec{p}_i,\vec{x}) + \tau^{\mathcal{P}}(\vec{p}_i) \Bigr\},
\end{equation}
where $\delta^\mathcal{M}(\vec{y},\vec{x})$ is the travel time along the geodesic path in $\bar{\Omega}$ connecting $\vec{x}$ and $\vec{y}$. Similarly,
\begin{equation}
\tau^{\mathcal{P}}(\vec{x}) = \min \Bigl\{ \delta^\mathcal{P}(\vec{p}_0,\vec{x}) + t_0, \min_{i=1,\ldots,M_t} \Bigl\{  \delta^\mathcal{P}(\vec{p}_i,\vec{x}) + \tau^{\mathcal{M}}(\vec{p}_i) \Bigr\} \Bigr\},
\end{equation}
where $\delta^\mathcal{P}(\vec{y},\vec{x})$ is the travel time in the Purkinje tree $\bar{\mathcal{P}}$.

The numerical algorithm is inspired by the above formula. Given $\tau_0^\mathcal{M}(\vec{x}) = +\infty$ we iterate, for $n\ge 0$,
\begin{equation}
\label{eq:eikoiter}
\left\{\begin{aligned}
\tau_{n+1}^{\mathcal{P}}(\vec{x}) &= \min \Bigl\{ \delta^\mathcal{P}(\vec{p}_0,\vec{x}) + t_0, \min_{i=1,\ldots,M_t} \Bigl\{ \delta^\mathcal{P}(\vec{p}_i,\vec{x}) + \tau^{\mathcal{M}}_n(\vec{p}_i) \Bigr\} \Bigr\}, \\
\tau^{\mathcal{M}}_{n+1}(\vec{x}) &= \min_{i=1,\ldots,M_t} \Bigl\{  \delta^\mathcal{M}(\vec{p}_i,\vec{x}) + \tau^{\mathcal{P}}_{n+1}(\vec{p}_i) \Bigr\}.
\end{aligned}\right.
\end{equation}
The algorithm readily generalizes to the case of two independent Purkinje trees for the left and right ventricle. We note that the myocardium may couple the trees: for instance, the right activation may occur much earlier than the left activation, thus myocardial activation in the left ventricle will enter into the left Purkinje tree and overshadow the intrinsic activation. This is the case of a left bundle branch block.

For the discretization in space, we consider a piecewise linear approximation $\mathcal{P}_h$ for $\mathcal{P}$ and a general conformal mesh $\Omega_h$ of $\Omega$. We approximate the conduction velocity as a piecewise-constant function on the elements of the mesh. In this way, the numerical solution of the eikonal equation on $\mathcal{P}_h$ is equivalent to the Dijkstra algorithm on the tree. In the myocardium, we use the Fast Iterative Method~\cite{fu_fast_2013} implemented on GPU~\cite{Pezzuto2017Fast}. For the coupling, the PMJs are not necessarily vertices of the myocardial mesh. In this way, we can avoid remeshing of the myocardial mesh or constraining the PMJs during the tree generation. Hence, we apply the following boundary conditions:
\begin{equation}
\tau^\mathcal{M}(\mathbf{x}_j) = \sqrt{\ten{D}^{-1}(\vec{p}_i)(\vec{x}_j-\vec{p}_i)\cdot(\vec{x}_j-\vec{p}_i)} + \tau^{\mathcal{P}}(\vec{p}_i),
\end{equation}
where $\mathbf{x}_j$ are the nodes of the element containing $\mathbf{p}_i$. That is, we solve exactly the eikonal equation within the element.

The surface ECG is computed via the lead field approach~\cite{Pezzuto2017Fast,potse2018scalable}. Given the lead fields $Z_\ell(\vec{x})\colon\Omega\to\mathbb{R}$, for $\ell=1,\ldots,12$, the $\ell$-lead of the surface ECG is as follows:
\begin{equation}
V_\ell(t) = \int_{\Omega} \mathbf{G}_i(\vec{x}) \nabla U\bigr(t-\tau^{\mathcal{M}}(\vec{x}) \bigr)\cdot \nabla Z_\ell(\vec{x})\,\mathrm{d}\vec{x},
\end{equation}
where $U(\xi)$ is a template action potential. In this work, $U(\xi)$ has been computed from a 1D cable simulation with the Ten Tusscher-Panfilov model~\cite{TenTusscher2004}.
We neglect the contribution of the Purkinje tree on the ECG. The lead fields $Z_\ell(\vec{x})$ are computed once in the torso domain by solving the bidomain equation~\cite{potse2018scalable}.

\subsection{Purkinje network estimation}

Using the Purkinje network generation algorithm and the forward ECG model, we expect to find possible representations of the Purkinje tree for a specific patient.
For each case, we have the 12-lead ECG time series, that we call reference ECG $\hat{\mathbf{V}}(t)$. Then the task is to find possible trees for the patient Purkinje network, such that the generated ECG $\mathbf{V}(t;\vec{\theta})$ matches the data $\hat{\mathbf{V}}(t)$ in a least-squares sense. We note that the generated ECG $\mathbf{V}(t;\vec{\theta})$ depends on geometric and electrophysiology parameters of the Purkinje network, which we summarize in a vector $\vec{\theta}$. As geometric parameters, we choose the left and right initial length $l_{\rm i}^{L}$, $l_{\rm i}^{R}$; and 2 fascicles lengths and fascicles angles for each ventricle $l_{\rm{F}i}^{\rm L}$, $l_{\rm{F}i}^{\rm R}$, $\alpha_{\rm{F}i}^{\rm L}$, $\alpha_{\rm{F}i}^{\rm R}$ with $i \in \{1,2\}$. We fix the remaining geometric parameters, selecting a branch length $l_{\rm b}$ of 8 mm, the number of segments $N_s$ to 8, the repulsion parameter $w$ to 0.1, a branch angle $\alpha_{\rm b}$ to 0.15 rad, and 20 branch generations \cite{SahliCostabal2015}. A sensitivity study showed that accurate reconstructions were possible even if these parameters where fixed, thus reducing the dimensionality of the problem. For electrophysiology parameters, we choose the conduction velocity of the Purkinje network and the delay between the activation of the left and right ventricle, which we call ``root time''. If this parameter is positive, it means that the right Purkinje tree activated later than the left, and vice versa. Considering both geometric and electrophysiology parameters, we need to estimate 12 parameters, thus $\vec{\theta} \in \mathbb{R}^{12}$. We define a plausible range for the parameters for the search, shown in Table~\ref{tab:param_range}.

\begin{table}[]
\centering
\caption{Geometric and electrophysiology parameters, with their symbols and search range. The $(\cdot)^{L}$ and $(\cdot)^{R}$ indicate the left and right ventricle parameters, respectively.}
\label{tab:param_range}
\begin{tabular}{@{}llll@{}}
\toprule
                                              & Parameter           & Symbol        & Range                    \\ \midrule
\multirow{4.}{*}{Geometric parameters}       & Initial length      & $l_{\rm i}^{L}$, $l_{\rm i}^{R}$  \vspace{5pt}  & 30 to 100 mm \\
                                              & Fascicles lengths   & $l_{\rm{F}1}^{\rm L}$, $l_{\rm{F}2}^{\rm L}$, $l_{\rm{F}1}^{\rm R}$, $l_{\rm{F}2}^{\rm R}$ \vspace{5pt} & 2 to 50 mm  \\
                                              & Fascicles angles    & $\alpha_{\rm{F}1}^{\rm L}$, $\alpha_{\rm{F}2}^{\rm L}$, $\alpha_{\rm{F}1}^{\rm R}$, $\alpha_{\rm{F}2}^{\rm R}$ \vspace{5pt} & -$\pi$/4 to 3$\pi$/4 rad \\ \cmidrule(r){1-1}
\multirow{2.5}{*}{Electrophysiology parameters} & Root time & RT \vspace{5pt}  & -75 to 50 ms             \\
                                                & Conduction velocity & CV   & 2 to 4 m/s               \\ \bottomrule
\end{tabular}
\end{table}

Thus, we have to find the parameters $\vec{\theta}_{\min}$ that minimize the difference between $\hat{\mathbf{V}}(t)$ and $\mathbf{V}(t;\vec{\theta})$.
We compute the mismatch by taking into the account only the QRS complex of the ECGs, which corresponds to the ventricular activation.
Further details on the pre-processing of the ECG are given in Section~\ref{sec:patdata}. The QRS onset of the simulated ECG may also vary, because the Purkinje networks vary and also because the Q deflection, which corresponds to the first epicardial breakthrough, is usually 10-15 ms late with respect to endocardial activation. Therefore, we use cross-correlation to find the best alignment between the signals. Then, as an optimization objective, we select the $L^2$ norm of the difference between the QRS complexes:
\begin{equation}
    \vec{y}(\vec{\theta}) = \frac{1}{T} \int_0^{T} \left(\hat{\mathbf{V}}(t) - \mathbf{V}(t;\vec{\theta})\right)^2 \mathrm{d}t,
    \label{eq:loss}
\end{equation}
where $T$ is the QRS duration.

Finally, we aim to solve the following parameter estimation problem:
\begin{equation}
    \vec{\theta}_{\min} = \argmin_{\vec{\theta}} y(\vec{\theta}).
    \label{eq:inverse}
\end{equation}

The summary of the steps is shown in Figure \ref{fig:outline}. We use Bayesian optimization for solving~\ref{eq:inverse}.

\subsection{Bayesian optimization}\label{subsection:bay_opt}

Bayesian optimization is an approach to efficiently minimize complex and costly functions, used in areas such as machine learning, physical systems, and field experiments \cite{blanchard2021bayesian,garnett2023bayesian}. This approach combines principles of Bayesian inference and optimization, to find a global optimal solution while minimizing the number of data required. There are two main aspects of the Bayesian optimization. The first one is the surrogate model, which approximates the objective function given known data points. The second is the acquisition function, which selects new points in the parameter space by balancing the exploration of regions with high uncertainty and the exploitation of areas with low objective function. 

In our application, for each patient, we begin by training a surrogate model with $N$ previously computed points $\{ \vec{\Theta},\vec{y}\} = \{(\vec{\theta}_i,y_i)_{i=1}^{N}\}$, where $y_i$ represents the mismatch between measured and generated ECGs, defined in equation~\eqref{eq:loss}, and $\vec{\theta}_i$ represents the parameters of the Purkinje network, as previously described.
Then, we construct our surrogate model by representing the relationship between $\vec{\theta}$ and $\vec{y}$ as
\begin{equation}
    \vec{y} = f(\vec{\theta}) + \varepsilon,
    \label{eqn:f}
\end{equation}
where $f$ is the latent function we want to infer and $\varepsilon$ is a noise that may corrupt the output.
For simplicity, we assume $\varepsilon$ as Gaussian and uncorrelated, $\varepsilon \sim \mathcal{N}(\vec{0}, \sigma_n^2\vec{I})$, where $\sigma_n^2$ is an unknown variance parameter that will be learned from the data. In the case where the relationship between $\vec{\theta}$ and $\vec{y}$ is noiseless (i.e. deterministic), a small level of noise is still considered for numerical reasons.
For the latent function $f$ we assume a zero-mean Gaussian process prior
\begin{equation}
    f(\vec{\theta}) \sim \mathcal{GP}\bigl(\vec{0},k(\vec{\theta},\vec{\theta}';\vec{\beta})\bigr),
\end{equation}
where $k(\vec{\theta},\vec{\theta}'; \vec{\beta})$ is the covariance kernel function, which represents our knowledge about the objective function behavior and depends on a set of hyper-parameters $\vec{\beta}$.
In each step of the optimization, with the known data $\{ \vec{\Theta},\vec{y}\}$ we train the model with the end of finding the hyper-parameters $\vec{\beta}$ that explain their underlying relationship.
Here we adopt the Automatic Relevance Determination (ARD) exponential kernel \cite{despotovic2020speech,kumar2021selection}
\begin{equation}
    k(\vec{\theta},\vec{\theta}';\vec{\beta}) = \eta^2 \cdot \exp(-r),
\end{equation}
where $\vec{\beta}=\{\eta, r_1, ..., r_n \}$, with $\eta$ the signal standard deviation and
\begin{equation}
    r = \sqrt{\sum_{i=1}^{n} \frac{(\theta_i-\theta'_i)^2}{r_i^2}},
\end{equation}
where $r_i$ is the length scale of the kernel function in the input dimension $i$, so with increasing $r_i$ the influence of the $i$-th input decreases.
Once the model is trained we can obtain the posterior mean $\mu(\vec{\theta})$ and variance $\sigma^2(\vec{\theta})$ at any point, which represent the predicted value of our surrogate model and its uncertainty, respectively \cite{rasmussen2006gaussian}.
With this information we make use of the acquisition function to determine $\vec{\theta}_{N+1}$, the next point to acquire and incorporate to our dataset. This function varies in the parameter space and depends on value of the prediction and its uncertainty, such that $a(\mu(\vec{\theta}), \sigma^2(\vec{\theta}))$. It is expected that the point where the acquisition function reaches its maximum will have minimum objective function value. Then, to select the next point where to acquire the data, we solve the following optimization problem:
\begin{equation}
    \vec{\theta}^{N+1} = \argmax_{\vec{\theta}} a(\mu(\vec{\theta}), \sigma^2(\vec{\theta}))
    \label{eq:acq}
\end{equation}

There are many options of acquisition functions, here we select expected improvement \cite{jones1998efficient}, one of the most widely-used in Bayesian optimization.
Every acquisition requires the solution of the forward ECG model to obtain the new pair $(\vec{\theta}_{N+1}, y_{N+1})$.
For each simulation presented in this work, first we use the forward ECG model to sample 250 points with Latin hypercube sampling, and then we perform the Bayesian optimization for 300 steps.

\subsection{Posterior distribution estimation}

After the optimization, the goal is to estimate the posterior distribution of the Purkinje network parameters. For this task, we use Approximate Bayesian Computation (ABC)~\cite{sunnaaker2013approximate}. ABC is a likelihood-free inference technique that only requires defining a prior distribution and a distance function. Algorithmically, this method is straightforward: it generates a sample from the prior distribution and it accepts it if the distance between the measurements and the simulated data is less than a given tolerance. This process is repeated until sufficient samples are accumulated. We take advantage of trained Gaussian process surrogate for the Bayesian optimization to create an informed prior distribution. Here, samples with a low predicted value of the loss function will be assigned a higher probability to enter the posterior distribution. We also take into account the predicted uncertainty of the Gaussian process and assign less prior probability to points with high predicted variance. With these two considerations, we define the prior probability density as:
\begin{equation}
    p(\vec{\theta}) = \frac{1}{\sqrt{2\pi\left(\sigma^2(\vec{\theta}) + \sigma^2_{\rm min}\right)}} \exp\left({-\frac{1}{2}\frac{(\mu(\vec{\theta}) - y_{\rm min})^2}{\left(\sigma^2(\vec{\theta}) + \sigma^2_{\rm min}\right)}}\right)
    \label{eq:prior}
\end{equation}
where $y_{\rm min}$ and $\sigma_{\rm min}^2$ correspond to the mean and variance prediction of the Gaussian point at the point $\vec{\theta}_{\rm min}$ found during the optimization. We note that even though the prior distribution resembles a Gaussian, it is not possible directly sample $\vec{\theta}$ from its definition, as the mean and variance depend on $\vec{\theta}$ through a Gaussian process.
Thus, we rely on
rejection sampling \cite{robert2013monte}, a well-known algorithm to obtain samples from a target distribution. To obtain samples from the prior distribution, we generate $N_s = 5.000.000$ uniformly distributed random samples of $\vec{\theta}$. For each point, we use the Gaussian process previously trained (with the 250 initial points plus the 300 points found by the optimization) to compute the predicted value $\mu(\vec{\theta})$ and variance $\sigma(\vec{\theta})^2$.
Now we compute the prior density of each point using equation~\eqref{eq:prior} and compare $p(\vec{\theta})$ to a random variable sampled from a uniform distribution $r \sim \mathcal{U}(0,p_{\rm max})$. If $p(\vec{\theta}) > r$, the sample is accepted, otherwise it is rejected. We define the maximum probability $p_{\rm max}$ to be the prior probability in equation~\eqref{eq:prior} evaluated at $\mu(\vec{\theta}) = y_{\rm min}$ and $\sigma(\vec{\theta})^2 = \min_{\vec{\theta} \in S}(\sigma^2(\vec{\theta}))$, that is the minimum variance across the $N_s$ samples of $\vec{\theta}$. We note that there are potentially higher values of $p(\vec{\theta})$. However, this approach led to a higher acceptance rate for samples of the prior distribution, which will be later confirmed to enter the posterior distribution.
Now our task is to select $N_f$ final samples that can be used as models of the patient's Purkinje network. According to the ABC methodology, we need to define a distance function between the data and the model to accept or reject samples from the prior distribution. First, we consider that there is some inherent variability in the ECG of each patient. To take into account these variations, we define a distribution $q(e;\vec{\theta})$ of errors $e$ that is calculated as the error of each of the $N_b$ beats that we have available for a patient and the ECG predicted by the model with parameters $\vec{\theta}$. Then, we define our distance function as the total variation distance  \cite{devroye2018total} between the distribution obtained with $\vec{\theta}_{\rm min}$ and any other point $\vec{\theta}$:
\begin{equation}
    D(\vec{\theta},\vec{\theta}_{\rm min}) = \frac{1}{2} \int_{0}^{\infty} \bigl| q(e;\vec{\theta}) - q(e;\vec{\theta}_{\rm min})\bigr| \,\mathrm{d}e. 
    \label{eq:TV}
\end{equation}
We note that we only have samples of $q$, thus we employ the kernel density estimation function implemented in \texttt{scikit-learn}~\cite{scikit-learn} with a Gaussian kernel to estimate the integral in equation~\eqref{eq:TV}. We use cross-validation to obtain the optimal bandwidth of the kernel.

With ABC algorithm completely specified, we proceed to estimate the posterior distribution. First, we sort the samples previously accepted by descending prior probability $p(\vec{\theta})$. Then, for each set of parameters $\vec{\theta}$ we compute the forward ECG model and evaluate its error distribution $q(e;\vec{\theta})$. Now, we compute the distance between the predicted ECG and the best ECG. We accept the sample $\vec{\theta}$ if $D(\vec{\theta},\vec{\theta}_{\rm min}) < 0.9$. We repeat this process until we accept $N_f = 30$ posterior samples. If we reject 50 consecutive samples, we retrain the Gaussian process surrogate including all the new points calculated and we repeat the prior and posterior sampling.

\subsection{Patient's data}
\label{sec:patdata}

The proposed model is retrospectively tested with the data from four heart failure patients, which includes a 12-lead ECG and a clinically indicated cardiac magnetic resonance (CMR) scan with late gadolinium-enhancement (LGE) for scar detection.
Clinical data and model construction has been previously described~\cite{Maffessanti2020scar,pezzuto2021reconstruction}.
The institutional review board approved the study protocol, and all patients gave written and oral informed consent for the investigation. The study was performed in compliance with the Declaration of Helsinki.

Patient anatomies (heart, torso, lungs, major blood masses) were semi-automatically reconstructed from the CMR images. Surfaces of relevant anatomical structures were constructed from their contours using Blender (The Blender Foundation). The outline of the ventricles was detected from short-axis cine images. Scarred tissue was inferred from LGE-MRI sequence by manual contour and alignment. The cardiac anatomy was eventually discretized to create 1-mm resolution 3D volumetric computational grid. Ventricular fiber orientation was assigned using a rule-based approach.

The 5-minute ECG was recorded using a clinically available ECG machine (CS200 excellence, Schiller AG) having a sampling rate of 1000/s and an amplitude resolution of 1.0 $\mu$V. We pre-processed the ECGs as follows. We detected all beats in the recording, using the Pan-Tompkins algorithm~\cite{pan1985real}; we aligned the beats using the R-wave peak; we applied a linear detrend to each beat to correct the baseline; we manually excluded extra-sistoles and poor signals; we estimated the mean and standard deviation for each lead. The target QRS complex was finally isolated. The QRS onset and duration were detected manually on the mean signal.


For all patients, the values for the intra-/extra-cellular conductivities are: $\sigma^i_l = \sigma^e_l = \SI{3.0}{\milli\siemens\per\cm}$, $\sigma^i_t = \SI{0.3}{\milli\siemens\per\cm}$, $\sigma^e_t = \SI{1.2}{\milli\siemens\per\cm}$, and $\alpha$ is a patient specific value \numrange{0.05}{0.1}.


\section{Results}
\label{sec:results}
We test our method with two experiments: a synthetic example and 4 datasets of real patients. We finalize by simulating conduction system pacing from resynchronization therapy in the 4 patients.

\subsection{Synthetic case}

\begin{figure}[t]
	\centering
	\includegraphics[width=\textwidth]{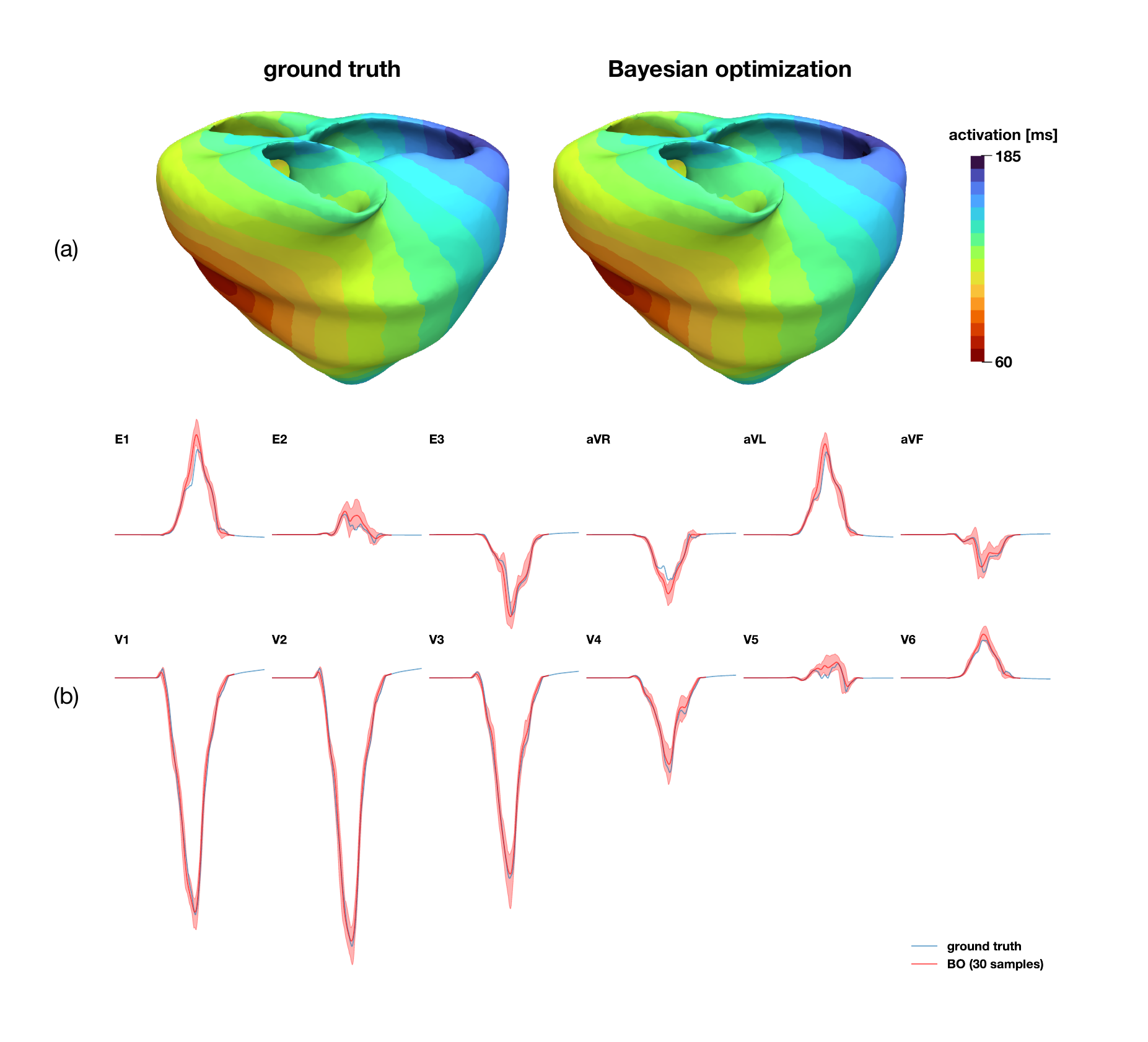}
	\caption{\textbf{Synthetic results.} (a) Activation time maps for the ground truth and the best network found by the optimization. (b) Reference ECG (in blue) and ECGs of the 30 networks found by the method (in red), with the mean in a solid red line and the area between the minimum and maximum values of each time step shaded.}
	\label{fig:synthetic_ECG}
\end{figure}

First we test the model in a synthetic example, where the reference ECG is obtained with the forward ECG model applied in a known Purkinje network.
To create the reference network we choose the parameters given by the minimum $\vec{\theta}_{\min}^{\text{pat1}}$ found by the optimization with the data of patient 1, using the method described in section \ref{subsection:bay_opt}. We select this approach in order to avoid the creation of a network with random parameters, which could lead to a non-realistic reference ECG.
Then, the parameters of the reference network are initial lengths of 35.93 mm (LV) and 79.86 mm (RV), fascicles lengths of 9.42 and 18.25 mm (LV) and 43.41 and 11.59 mm (RV), fascicles angles of 1.44 and 2.36 rad (LV) and 2.36 and 2.36 rad (RV), a root time of 75 ms in the left ventricle and a conduction velocity of 2 m/s, where LV and RV indicate the parameters of the left and right ventricles, respectively.
With these network parameters and the forward ECG model we construct the reference ECG and then we perform the optimization and the steps previously described.

\begin{figure}[t]
	\centering
	\includegraphics[width=1.\textwidth]{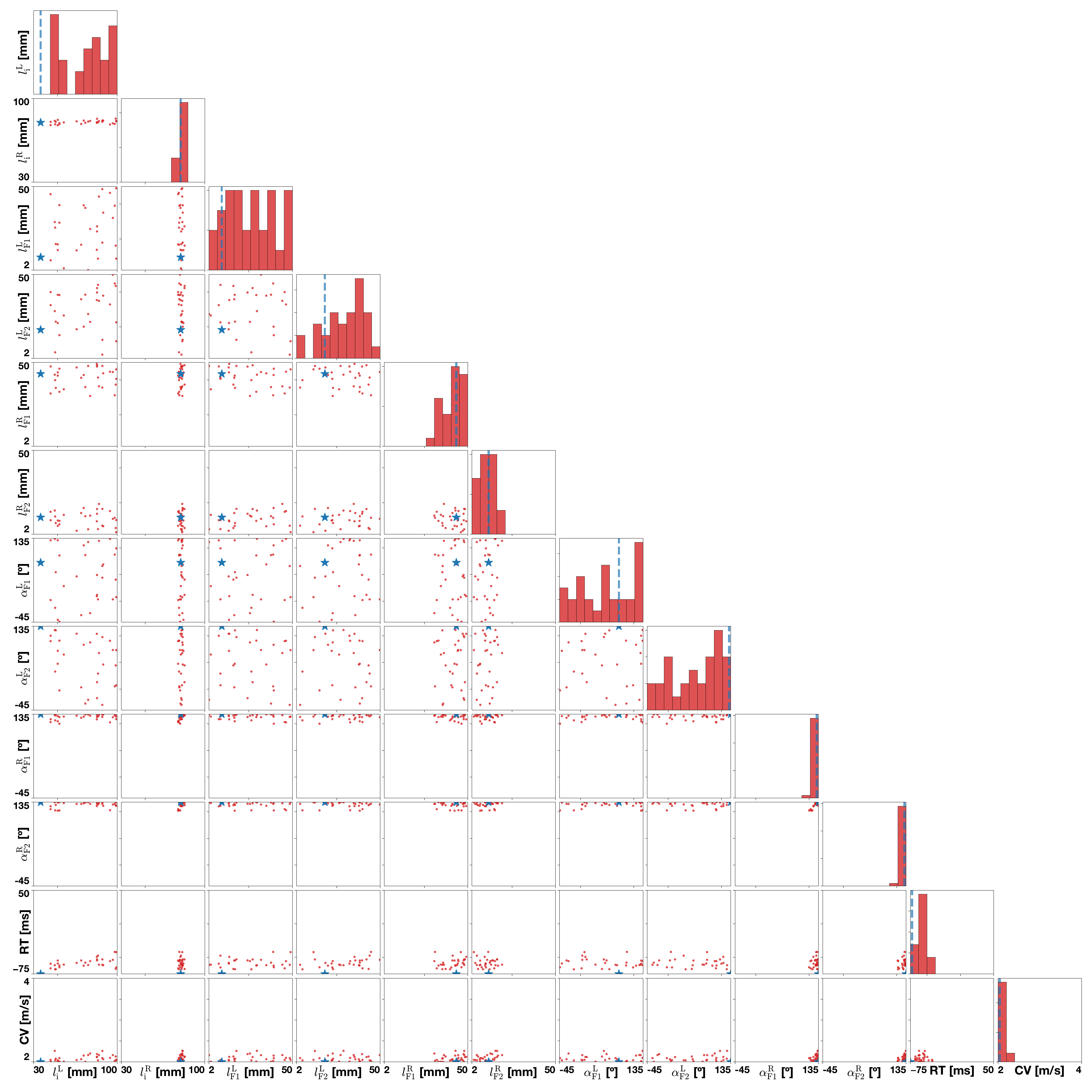}
	\caption{\textbf{Synthetic results.} The pair plot shows the distribution of the parameters of the reference network (in blue) and the inferred networks (in red).}
	\label{fig:synthetic_pairplot}
\end{figure}

\begin{figure}[t]
	\centering
	\includegraphics[width=1.\textwidth]{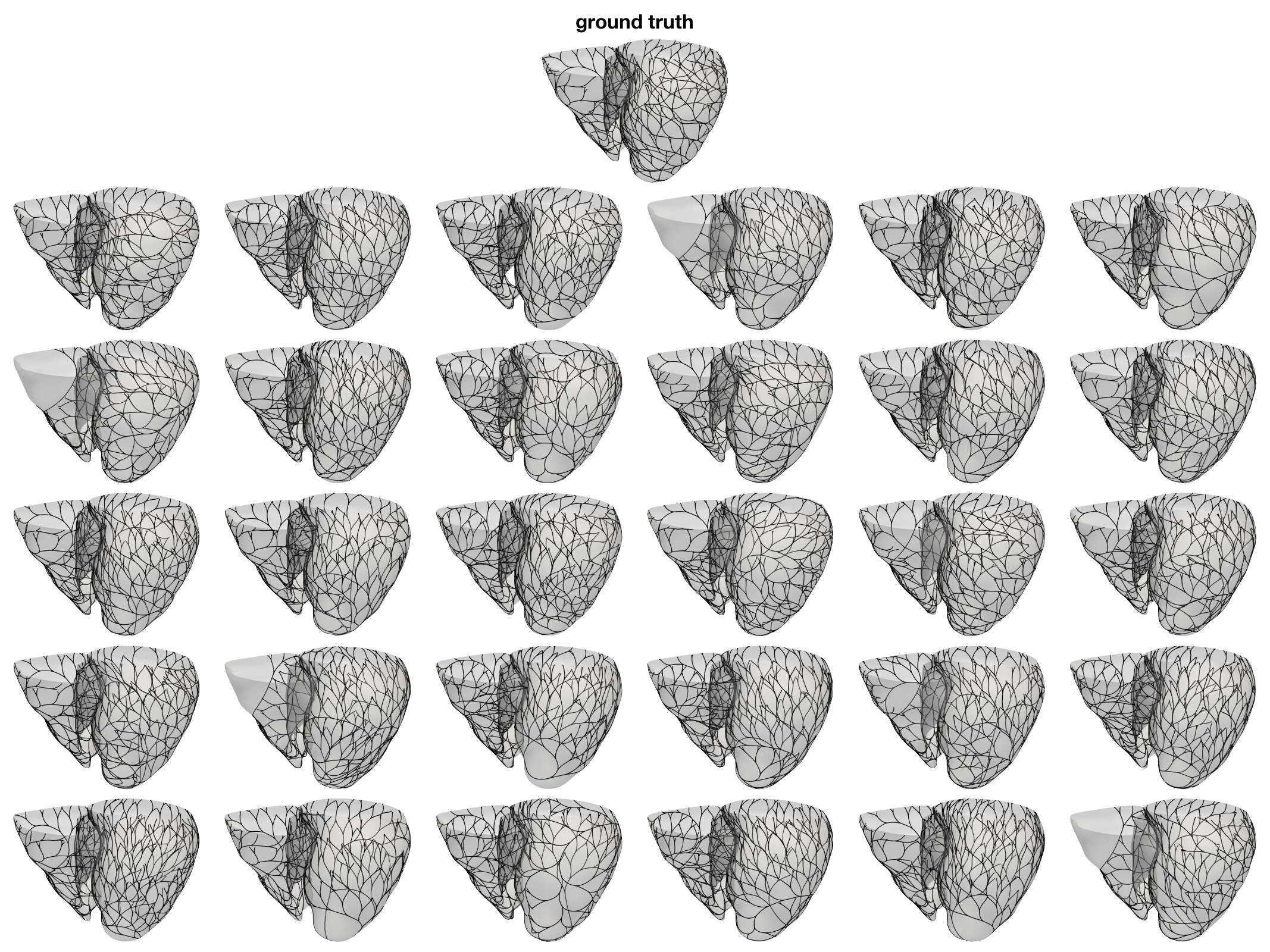}
	\caption{Inferred Purkinje trees for the synthetic example.}
	\label{fig:synthetic_tress}
\end{figure}

After running the entire pipeline, including the posterior distribution estimation, we obtain the results of the synthetic case in terms of activation time maps and predicted ECGs, shown in Figure \ref{fig:synthetic_ECG}. The distribution of the parameters of the networks selected is shown in the pair plot of Figure \ref{fig:synthetic_pairplot}.
The activation time maps of the reference network and the best point found by the optimization have very close pattern and values.
Regarding the 30 inferred networks, the predicted ECGs are similar to the reference ECG, with similar curves in all leads.
Additionally, the pair plot shows that the parameters of the right ventricle (initial length, fascicles lengths, fascicles angles) along with the root time and conduction velocity are concentrated around the reference values, while the other parameters are evenly distributed in their respective ranges. The different distributions of the parameters show that some of them can be easily identified, while other cannot. The lack of identifiability of the parameters of the left ventricle has a physiological explanation: the simulated patient has a left bundle branck block (root time of 75 ms in the left ventricle) and the activation of the left ventricle is coming directly from the right ventricle. This effect can be clearly observed in Figure~\ref{fig:synthetic_ECG}a, where the earliest activation occurs on the right ventricle and propagates towards the left ventricle. Thus, there is little effect of the left Purkinje tree on the ECG, and multiple combinations of parameters will explain the data. 
The variety of inferred Purkinje trees can be seen Figure \ref{fig:synthetic_tress}, where we present the reference network along with the 30 selected networks, which show how different geometries may produce similar ECGs.


\subsection{Patient data}
\begin{figure}[t]
	\centering
	\includegraphics[width=0.65\textwidth]{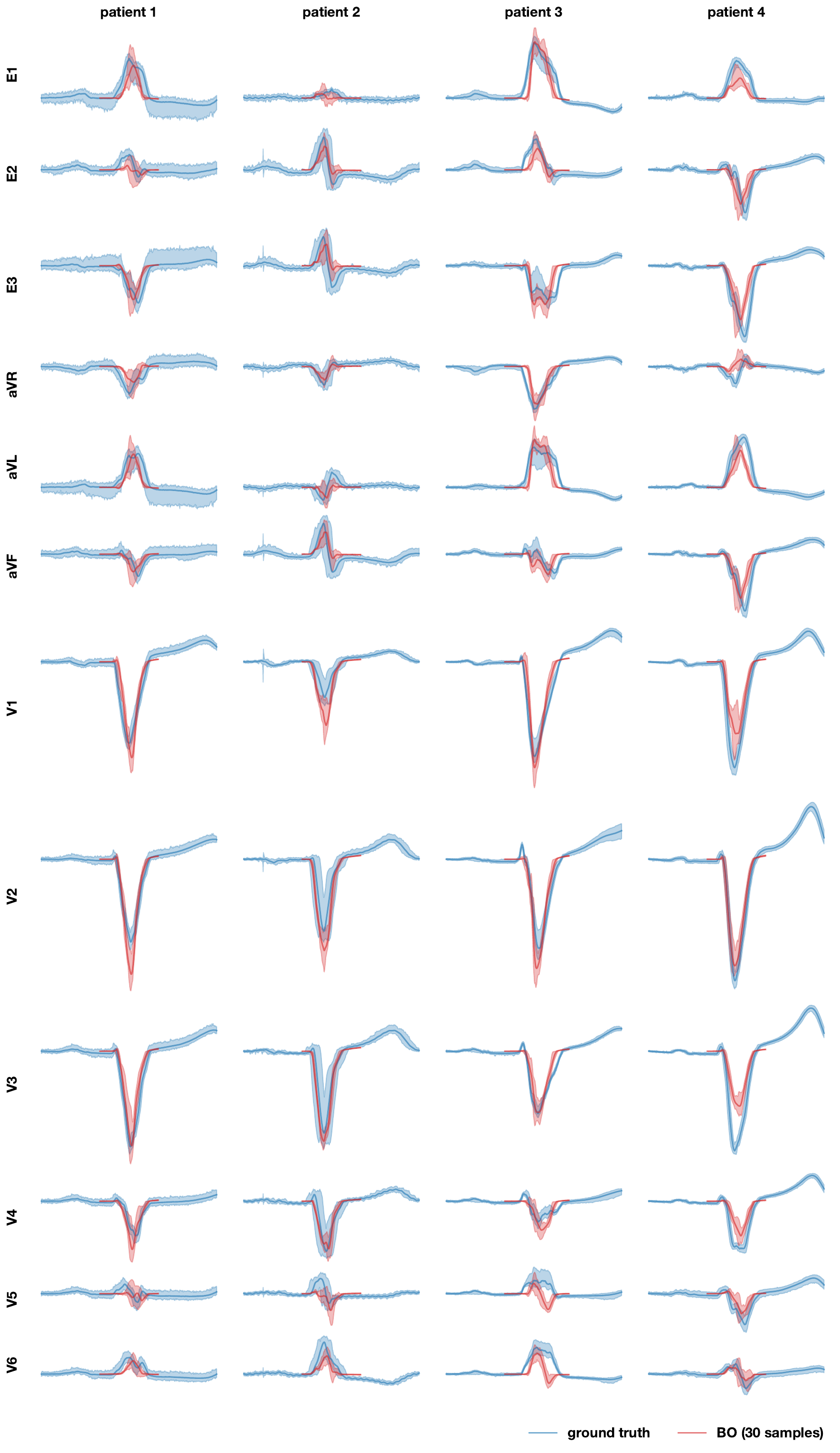}
	\caption{\textbf{ECGs for the four patients analyzed.} In blue, the reference ECG, with the area shaded between the minimum and maximum value of the signal of every beat. In red, the predicted ECGs of the 30 selected samples, with the mean in a solid red line and the area between the minimum and maximum value of each time step shaded.}
	\label{fig:patient_ecgs}
\end{figure}

Now we use the proposed model to find Purkinje networks that describe the ECG of each patient.
The optimization for patients 1, 2, 3, and 4 found minima with MSE 0.433, 0.348, 0.564, and 0.937, respectively.
Then, from the predicted ECGs shown in Figure \ref{fig:patient_ecgs}, it can be observed that the model correctly captures the behavior of each patient's ECG.
Moreover, the variability in the predicted ECGs is in the same scale as the inherent variability of the reference ECGs, depicted by the blue band.
However, the results show that the model struggles to capture  some of the low-amplitude leads, such as V5 and V6.

\begin{figure}[t]
	\centering
	\includegraphics[width=1.\textwidth]{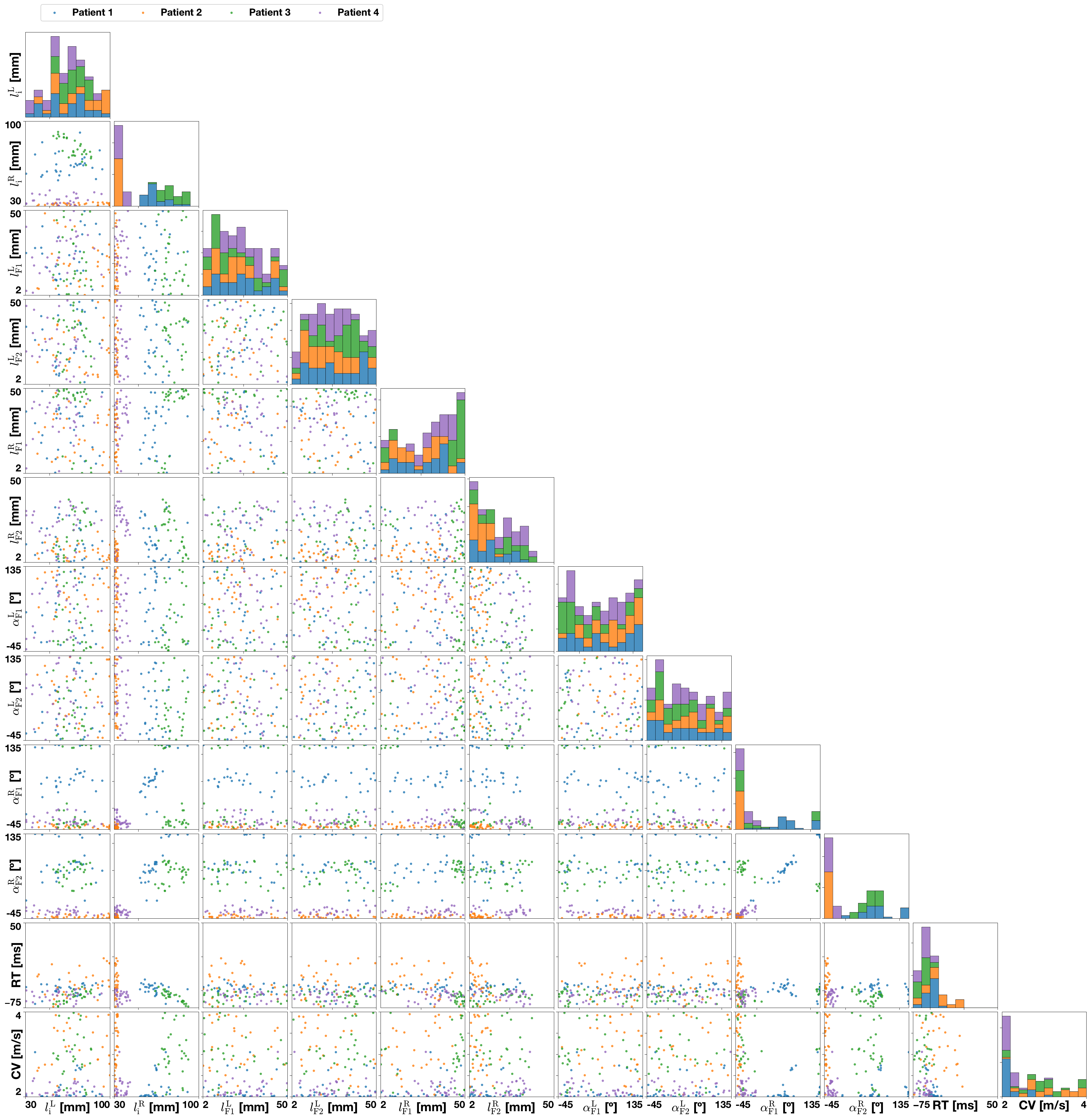}
	\caption{\textbf{Identified parameters for 4 patients.} pair plot with the parameters of the 30 networks selected for each patient.}
	\label{fig:patient_pairplot}
\end{figure}

The pair plot of Figure \ref{fig:patient_pairplot} shows the distribution of the parameters of the selected networks of each patient.
Note that some parameters are evenly distributed in their respective ranges, as the fascicles lengths in the left ventricle, which indicates that their values do not have an important effect in the ECG, since regardless their values it can be obtained an ECG similar to the reference.
In contrast, other variables are concentrated in certain regions. For example, the root time tends to have negative values, especially for patients 3 and 4, indicating a later activation of the left ventricle that may be caused by left bundle branch block.
For some variables the distribution depends on the patient, for example, for the conduction velocity the patients 1 and 4 (in blue and purple, respectively) are concentrated in the lower values, while the patients 2 and 3 (in orange and green) present values throughout the range, reflecting the characteristics of each patient.
Considering this, the parameters that can be identified from the data highly depend on the patient. For instance, patient 3 shows a concentrated distribution for the first fascicle length of the right ventricle, while patients 2 and 4 show a similar behavior for initial length of the right ventricle and the right fascicle angles.

\subsection{Conduction system pacing simulation}

\begin{figure}
	\centering
	\includegraphics[width=1.\textwidth]{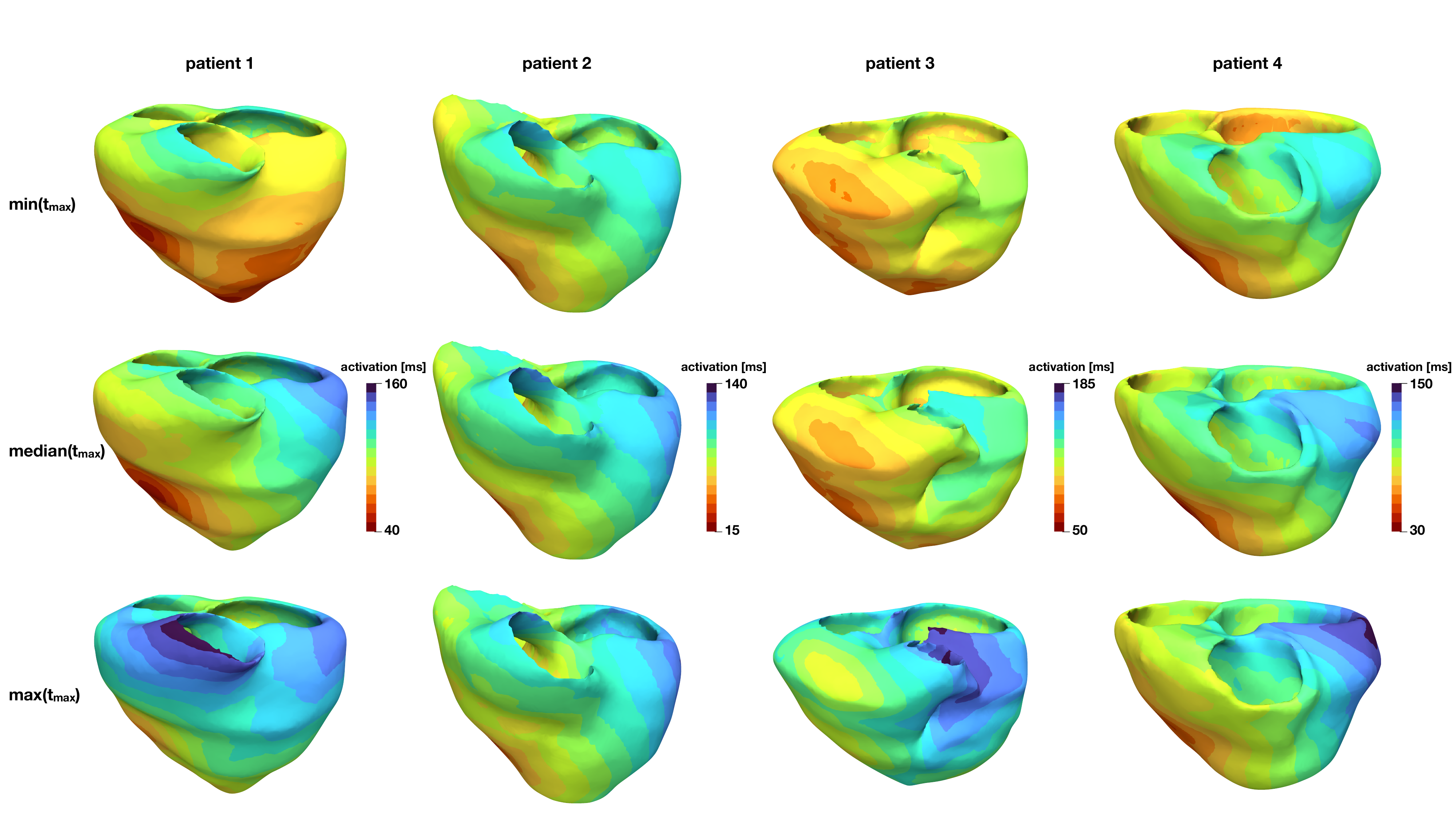}
	\caption{\textbf{Activation time maps for the pacemaker simulation.} For each patient, the maps are sorted according to their maximum activation time, but only the results with the minimum, the median and the maximum of these values are shown.}
	\label{fig:pacemaker_maps}
\end{figure}

\begin{figure}
	\centering
	\includegraphics[width=0.7\textwidth]{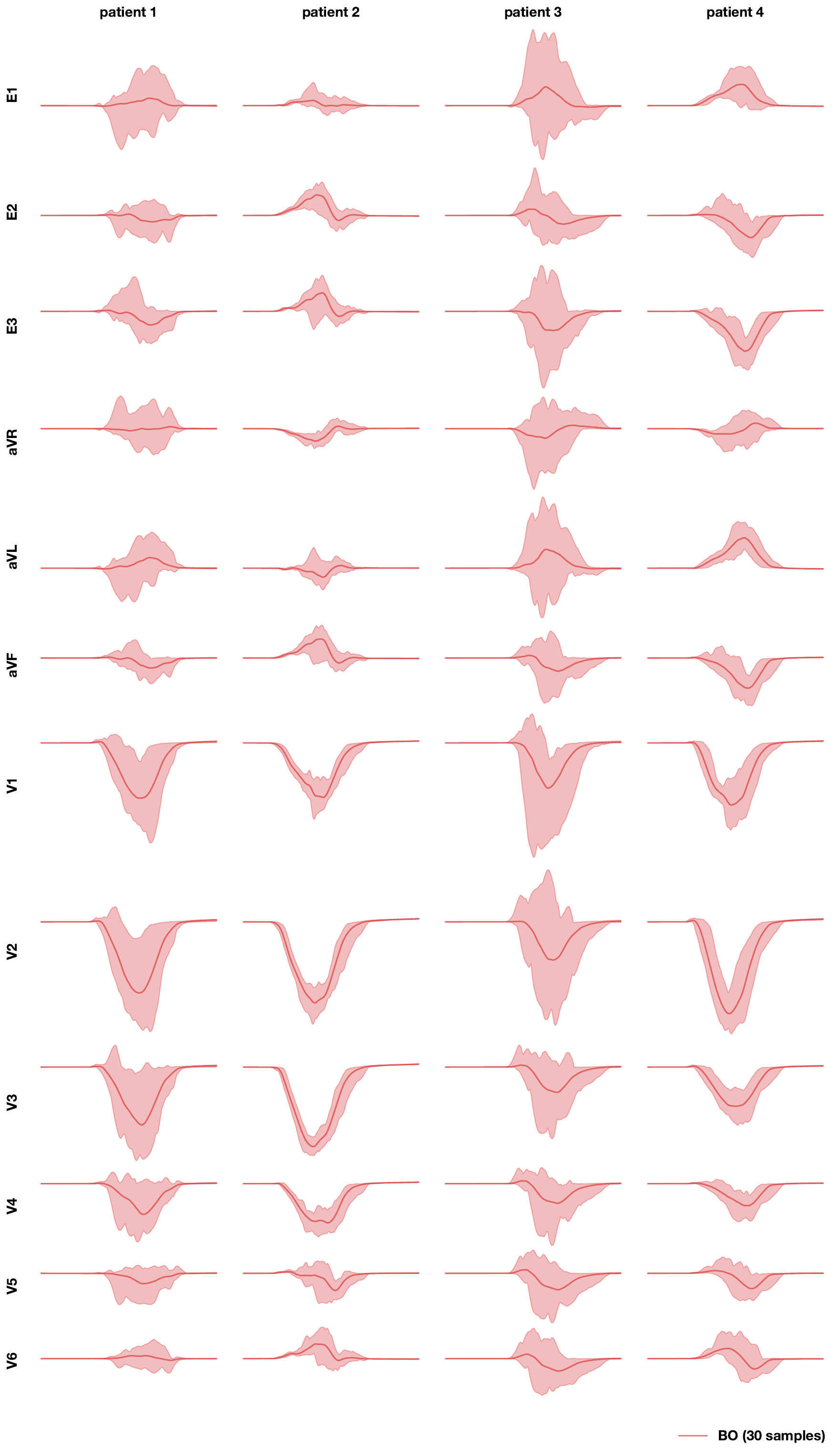}
	\caption{\textbf{ECGs of the pacemaker simulation.} In red, the predicted ECGs of the 30 selected samples of each patient, with the mean in a solid red line and the area between the minimum and maximum value of each time step shaded.}
	\label{fig:pacemaker_ecgs}
\end{figure}

The result of the proposed method is a set of Purkinje networks that can be used to predict a patient ECG under certain conditions or treatments, as a proof of concept here we simulate cardiac resynchronization therapy \cite{verzaal2022synchronization}. In particular, we attempt to recreate His-Purkinje conduction system pacing, where the pacemaker lead directly activates the Purkinje network to synchronize both ventricles \cite{vijayaraman2023his}. 
In order to simulate conduction system pacing, we take the selected networks of each patient and set the root time to 0 ms, synchronizing the activation of the left and right trees.
Figure~\ref{fig:pacemaker_maps} shows the activation time maps for the minimum, median and maximum largest activation time for each patient. We see that for patient 2, the activation maps show little variation, always displaying a right to left activation. Patient 4 shows a larger variation, but the earliest activation is located in the right ventricle in all cases. On the other hand, patients 1 and 3 show a larger variation in the activation maps. This is reflected in Figure \ref{fig:pacemaker_ecgs}, which shows the predicted ECGs for this simulation. Patients 1 and 3 show a much larger variability in the ECGs than patients 2 and 4. These results highlight the importance of taking a probabilistic approach, such that uncertainties in the predictions can be quantified. 

\section{Discussion}
\label{sec:discussion}

We have presented a novel method to identify the Purkinje network from the ECG.
This tool combines advanced machine learning techniques, with advanced cardiac electrophysiology modeling and statistical tools. We have developed a novel algorithm, based on the idea of harmonic maps, to generate the Purkinje network that is more efficient and stable than previous attempts. We also propose an approach to couple the Purkinje network with the myocardium efficiently. Finally, we design a custom algorithm combining Bayesian optimization with approximate Bayesian computation to infer a distribution of Purkinje networks that can explain the data. This point is particularly important in clinical setting~\cite{quaglino2018fast}, where the uncertainty are large and making a decision based on a certain but incorrect model can have dangerous consequences. We have extensively tested our methodology in synthetic examples and patient's data.

The synthetic example showcases that our pipeline is able to recover a distribution of parameters of the ground truth, serving as a numerical validation. More importantly, it highlights that recovering the Purkinje network solely from the ECG is an ill-posed problem with manifold solutions, as different combinations of parameters resulted in very similar ECGs. Moreover, portions of the Purkinje tree can become undetectable from the ECG, especially in the case of bundle branch block. Our method can detect this issue, which results in a large spread in the posterior distribution of the parameters.

In our tests with patient's data, we show that our methodology can recover the specific ECGs of each case. The posterior distribution of the parameters shows that there is a large variability on the parameters of the Purkinje network for the different patients. The ECGs of the patients also show some variability, which may explain the difference in parameters. The results of the posterior distribution also show that the parameters that can be identified depend on the patient, thus it would not be easy to eliminate a priori some parameters from the analysis. Nonetheless, a large uncertainty in some of the parameters does not necessarily imply a large uncertainty in the activation map. Therefore, the ECG-based posterior distribution could be a valuable clinical tool, for instance in the detection of the latest activated area or potential targets for lead placement~\cite{ellenbogen2023evolving}.

The cardiac resynchronization therapy study highlights a potential clinical application of our method. Here, we can use the inferred network to predict the activation patterns and ECGs after the application of a conduction system pacing. Coupled with a mechanical model, this tool could be used to quantify changes in ejection fraction for a given treatment \cite{lee2018computational}. But more importantly, this application shows the importance of estimating a distribution of Purkinje networks rather than a point estimate. In this way, we can estimate if we have actually learn something from the data and if our predictions are trustworthy. We saw in 2 out of 4 patients that the uncertainties in the predicted ECGs were large, which caused mainly by the large uncertainties in the parameters of the left Purkinje tree, which is not contributing to the surface ECG. Thus, when the left Purkinje tree is activated with the pacing, the resulting activation patterns are highly variable.

Our work presents some limitations. We use a simplified and idealized representation of the Purkinje network. However, from a modeling perspective, it is complex enough to represent the ECG of the patients that we studied. We even fixed some of the parameters such as the branch length, angle and the repulsive parameter, because we observed that we could still fit the ECG with these simplifications. In this sense, we do not claim that we can reconstruct the actual geometry of the Purkinje network, but a functional model of it. 
Another limitation is that creating the heart and torso model can be time consuming, thus it limits the clinical usability of our approach. However, deep learning techniques have greatly accelerated this process, making it possible to go from image directly to the model \cite{kong2022learning}. Regarding the computational cost to perform the inference of the Purkinje network, it took around 15 hours per patient, for the entire process. Since this is a process that can be done offline after the ECG and the images are acquired, it does not limit the clinical applicability. Nonetheless, the current implementation is not optimized and the computational time could be reduced by several hours. For instance, the Purkinje network generator is not optimized and written in plain Python. A compiled version would be 1-2 order of magnitude faster. This process currently takes much longer than solving the Eikonal equation on the entire heart, which is based on a code that is highly optimized for GPU~\cite{Pezzuto2017Fast}. Furthermore, we could run multiple models in parallel for the initial training set of the Gaussian process and for the approximate Bayesian computation.
Another limitation is that we have used an Eikonal model for ventricular and Purkinje activation. Importantly, we do not account for a realistic Purkinje-muscle junction. However, studies have shown that error introduced by the eikonal approximation is fairly small~\cite{neic_efficient_2017} and can be further reducted with an eikonal-diffusion model~\cite{GanderEikoDiff23}. Finally, 
we have only calibrated the QRS complex, excluding the T-wave. Although this is simplification, we believe is the only way to keep the problem tractable, as the morphology of the T-wave is due to spatial heterogeneities of the action potential duration, which is difficult to capture with an eikonal model. Nonetheless, once we have estimated the Purkinje network, we could regionally modify the model for the action potential to fit the T-wave as well~\cite{verzaal2022synchronization}. We plan to do this in the future.

Overall, we believe the presented methodology provides a solid first step to learn the Purkinje network from the ECG with quantified uncertainties. We hope it will improve the way that digital twins for cardiac electrolophysiology are created, leading into more precise diagnosis and effective treatment for patients.

\section{Acknowledgments}

This work was funded by ANID – Millennium Science Initiative Program – ICN2021\_004 and National Center for Artificial Intelligence CENIA FB210017, Basal ANID to FSC. FSC and FAB and also acknowledges the support of the project FONDECYT-Iniciaci\'on 11220816 and ERAPERMED-134 from ANID. We also acknowledge the CSCS-Swiss National Supercomputing Centre (project no.~s1074).



\bibliographystyle{elsarticle-harv}
\bibliography{litra}




\end{document}